\documentclass[letter]{spie}  

\usepackage{amsmath,amsfonts,amssymb}
\usepackage{graphicx}
\usepackage[colorlinks=true, allcolors=blue]{hyperref}
\usepackage{nicefrac}
\usepackage{subfigure}
\usepackage{booktabs}
\usepackage{multirow} 
\usepackage{hyperref}

\usepackage{url}

\usepackage{color}

\usepackage{color}

\title{Multisensor Data Fusion for Automatized Insect Monitoring (KInsecta)}

\author[a]{Martin Tschaikner}
\author[a]{Danja Brandt}
\author[a]{Henning Schmidt}
\author[a]{Felix Bie\ss mann}
\author[a]{Teodor Chiaburu}
\author[b]{Ilona Schrimpf}
\author[b]{Thomas Schrimpf}
\author[b]{Alexandra Stadel}
\author[a]{Frank Hau\ss er}
\author[a]{Ingeborg Beckers}
\affil[a]{Berliner Hochschule für Technik, Luxemburger Str.10, 13353 Berlin, Germany}
\affil[b]{UBZ Listhof, Friedrich-List-Hof 1, 72770 Reutlingen, Germany}

\authorinfo{Send correspondence to Frank Hau\ss er or Ingeborg Beckers \\ E-mail: hausser@bht-berlin.de, beckers@bht-berlin.de}

\begin{document} 
\maketitle

\begin{abstract}
Insect populations are declining globally, making systematic monitoring essential for conservation. Most classical methods involve death traps and counter insect conservation. This paper presents a multisensor approach that uses AI-based data fusion for insect classification. The system is designed as low-cost setup and consists of a camera module and an optical wing beat sensor as well as environmental sensors to measure temperature, irradiance or daytime as prior information. The system has been tested in the laboratory and in the field. First tests on a small very unbalanced data set with 7 species show promising results for species classification. The multisensor system will support biodiversity and agriculture studies. 
\end{abstract}

\keywords{data fusion, insect monitoring, optical wingbeat sensor, machine learning}

\section{INTRODUCTION}
\label{sec:intro}  

Insects play a crucial role in various ecological and economic interactions with their environment, e.g. as a food source for various animals and due to their pollination services. They also play an important part in composting for soil fertility and in cleaning water. However, the number of insect species and individuals is in sharp decline worldwide. In Germany, for example, the Krefeld study recorded a 75\% decline in insect biomass between 1989 and 2015 \cite{Hallmann2017} \cite{sanchez2019worldwide}. The reasons for this decline are manifold. Systematic monitoring of population, occurrence and distribution is therefore of great importance.

Conventional insect monitoring systems usually use death traps to measure the absolute biomass of dried insects. Only a few species are classified at the species level. In this paper, a live monitoring system is presented that can complement and extend conventional long-term monitoring. Especially by including citizen scientists, a data set can be obtained that is important for systematic correlation studies. That way, blind spots on the monitoring map can be prevented. 
We developed a standardized multisensor system that is easy to use, combined with a Web application for data collection and community networking. To bring monitoring to the public, an automated system based on AI algorithms is used, multiplying the valuable and necessary expert knowledge that is essential for serious statistics in scientific studies. The system counts and classifies insects according to the GBIF taxonomic system (order, family, genus, species). The main focus in the development of the system is the reproducibility of the data. Second, a cost-efficient selection of the components was important. Since some species are rare and the data base is strongly unbalanced, good data are essential for a robust automatic classification.
Current research in this area focuses on three partially overlapping use cases: Unstandardized mobile and open-source applications for citizen science  \cite{van2018inaturalist}, monitoring for pest control and identification of beneficial insects in agriculture \cite{lima2020automatic, wu2019ip102}, and biodiversity monitoring \cite{vanKlink2022}.

This paper presents a system that catches up with traditional monitoring by
providing a live monitoring system, which can be connected to various established insect traps but may also allow for modified trap systems. 

Automated monitoring uses a variety of technologies to detect, count, and distinguish insect species. Previous monitoring approaches have relied mainly on capturing image data. Others used wingbeat frequencies measured by acoustic sensors, multispectral analysis of reflected light, capacitance changes, and optoacoustic sensors \cite{Mankin2021, Rigakis2019, Potamitis2018, Wang2020}. For imaging, insects must be at rest for the camera to take well-focused snapshots. Those studies that perform training of machine learning algorithms based on data from live insects, rather than image data from museums and other collections of dead insects, use yellow fields or light traps to image insects at rest \cite{Sittinger2023}.

The multisensory design of the monitoring system is presented below. It consists of a camera system, a wing beat detector, and environmental sensors to measure temperature, humidity, and spectral irradiance. We use supervised learning for the training of the neural network on the basis of experts labeling a ground truth. The classification is done for hierarchical according to GBIF. In a prove of concept, it is demonstrated that combining different sensory signals via timestamps the prediction accuracy is improved compared to classifying on the base of the response of individual sensors only.
\section{Multisensor System}

\begin{figure} [ht]
   \begin{center}
   \includegraphics[width=0.8\textwidth]{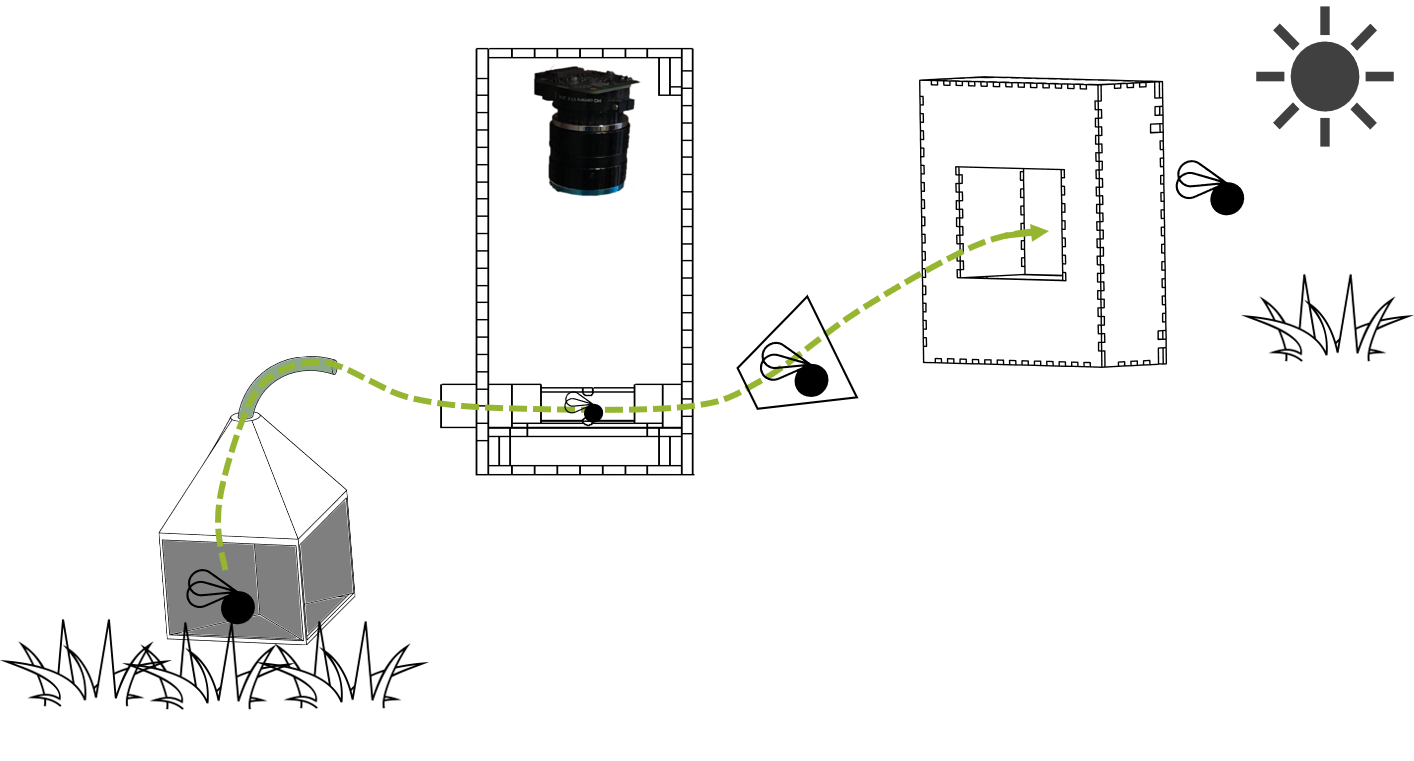}
   \end{center}
   \caption[wb] 
   { \label{fig:Systemprinciple} Principle of Multisensorsystem including the path of an insects exemplarily}
   \end{figure}

The multisensor system of the present work consists of a camera system with a resolving power of 10 micrometers, an optoacoustic wing beat sensor, environmental sensors and a real-time clock to correlate the data. The center of the sensor system is a Raspberry Pi4 minicomputer which is used for controlling and data processing. The insects first enter a camera system from a user-definable trap, from where they leave the system alive after passing the wingbeat sensor (fig.\ref{fig:Systemprinciple}).

The whole setup is open source and the components are chosen to be cost efficient according to the performance requirements. The instructions can be found on Gitlab (\url{https://gitlab.com/kinsecta/sensorik_dev/sensorcluster}). The data can be exchanged via a WebApp. There is also a forum for networking among participating monitoring professional scientists and Citizen Scientists. To increase the accuracy of the classification, the following environmental data are collected to make probability predictions about the occurrence of specific species. These include measurements of humidity, pressure and temperature using and Adafruit BME280 I2C sensor and the photometric brightness with an Adafruit BH1750. 

The spectral distribution of light intensity, measured with a Adafruit AS7341 10-Channel Light sensor, can be used to indirectly infer the cloudage. Hyperspectral data are detected for the wavelength channels 415nm, 445nm, 480nm, 515nm, 550nm, 590nm, 660nm, 690nm with bandwidth of $\pm$ 10nm respectively and an infrared sensor sensitive in the range of (850-1050)nm. An additional clear sensors measures the complete spectrum. Due to scattering by water droplets, the spectral distribution changes at different degrees of cloud coverage. A real time clock (Adafruit PCF8523) ensures the co-registration of data. We preferred commercially available sensors that communicate via I2C, that have a STEMMA QT interface and hence can be connected in series.

\subsection{Camera System}
\label{ss:camera}

Classification of insects based on images considers features such as color, contrast, patterning, size, and shape relationships. Entomological features such as pubescence and wing vascular pattern can be decisive for certain species. The developed system is standardized for homogeneous illumination, motion blur suppression, resolution, and depth of field, which are critical for accurate classification. Homogeneous diffuse illumination and resolution adjusted to the requirement of insect differentiation are crucial for classification. For example, background shadows make segmentation difficult. For this reason, diffusing PMMA glass provides uniform lightning from both lateral directions. With an effective aperture of f/8, a resolution of 10 micrometers is achieved, which is e.g. sufficient for differentiating vein segments in the wings of insects and also to identify single hairs. At the same time, a sufficient basis of depth of field is achieved in the so-called insect arena. The arena area measures 60x45x20 $\mathrm{mm}^{3}$.

A custom circuit board controls the lighting, which starts as soon as the insects pass a light barrier in the center of the arena. Two light barriers are integrated at the top and the bottom of the arena respectively, using two photoelectric lock-in amplifiers (IS471FE Sharp Microlectronics) to trigger the homogeneous LED illumination. A 500-microsecond flash with an exposure time of 23.5 milliseconds simulates a global shutter. Thus, the motion blur from the CMOS rolling shutter is minimized. The camera system is presented in detail in a previous paper from Brandt et al \cite{Brandt2023}.
Fluttering Wings and moving limbs show some motion blur, but the fuselage remains unblurred and sharp. The self-designed board is connected to a Raspberry Pi 4, which drives an HQ camera with a 10-MP teleobjective and stores an H264 video stream in a ring buffer. Simultaneously, with the triggering via the photoelectric sensor, three png-images are automatically extracted from the video stream selected via brightness. They are used for classification. The video sequence is also stored. A Python-based Bokeh-Application is running on a local server for controlling the sensors, data processing and storing information on SD card.  

\begin{figure} [ht]
   \begin{center}
   \includegraphics[width=0.4\textwidth]{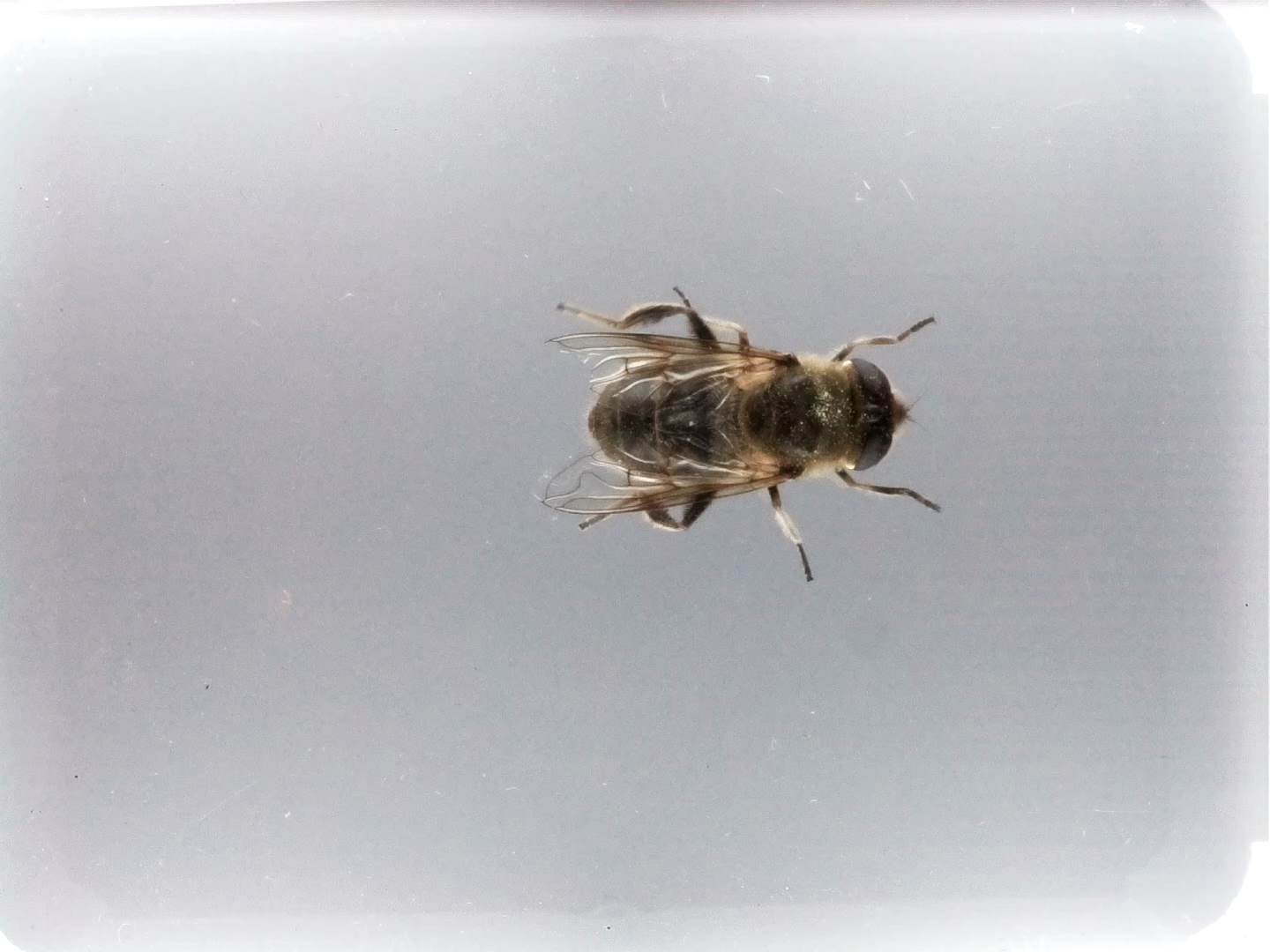}
   \end{center}
   \caption[cameraImage] 
   { \label{fig:cameraImage} Image taken by the camera system: An insect, species Eristalis tenax, is passing the arena.}
   \end{figure}

\subsection{Wingbeat Sensor}
\label{ss:wingbeat}

\begin{figure} [ht]
   \begin{center}
   \includegraphics[width=0.65\textwidth]{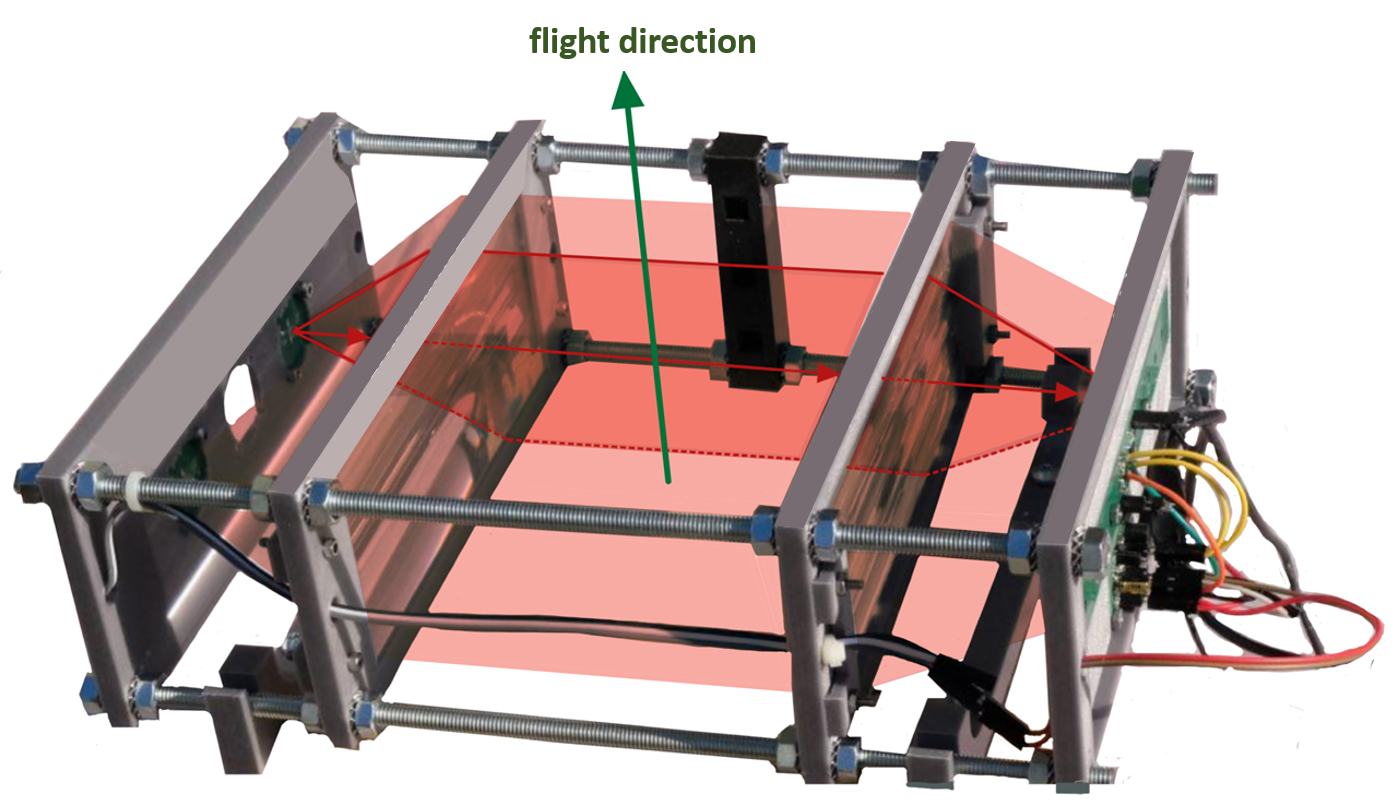}
   \end{center}
   \caption[wb] 
   { \label{fig:wb} Wingbeat sensor including Emitter, Receiver and Fresnel Lenses}
   \end{figure}

An optoacoustic sensor records the variation of infrared light intensity to record the wingbeat frequency of insects. 
As soon as an insect passes the active zone of the sensor, the light intensity decreases due to shading by the insect's body. The signal is modulated by the wingbeat frequency (fig.\ref{fig:wb}).

In contrast to acoustic measurements, here the wingbeat signals are detected free of background noise. The method based on an idea by Potamitis \cite{Potamitis2016,Potamitis2018}, is adapted for multisensor monitoring and allows for co-relation with camera images and other measurements via timestamp. The challenge in insect monitoring is the classification of diverse insects with a large variance in different wingbeat frequencies and flight velocities. The range of wingbeats can vary greatly depending on species, size, the physical condition of the insect, and temperature. But also the flight behavior like hovering, fast flight or navigation influences the frequency. The design of our wingbeat sensor focuses on the typical frequencies between 200-600 Hz, so that at reasonable flight speeds between 2 and 30 km/h the minimum active path through the sensor must be sufficient to measure 2-50 wing beats, which requires an active length of 6 cm. For this reason, we use Fresnel lenses that provide homogeneous illumination over the length along the direction of flight path. The sensor area is (100 x 100) mm$^{2}$. Light of two IR LEDs (YSMY12940, light intensity 16mW/sr, wavelength 940nm and an aperture angle of $\pm$ 40° FWHM) is collimated by cylindrical Fresnel lenses (f=50mm) and focused behind the active area onto 12 highly sensitive PIN photo diodes (QSB34GR, aperture angle $\pm$ 60° FWHM, rise and fall times are 50ns), ensuring a sampling rate of up to 2.5MHz. The wing beat frequencies can thus be sampled in detail.

\begin{figure} [ht]
   \begin{center}
   \includegraphics[width=0.49\textwidth]{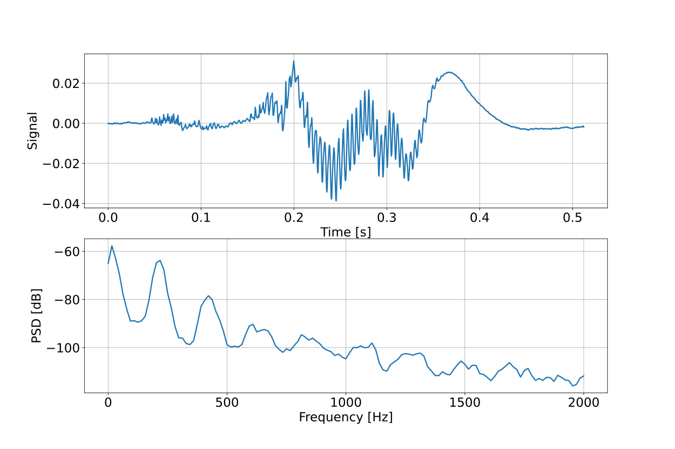}
   \includegraphics[width=0.49\textwidth]{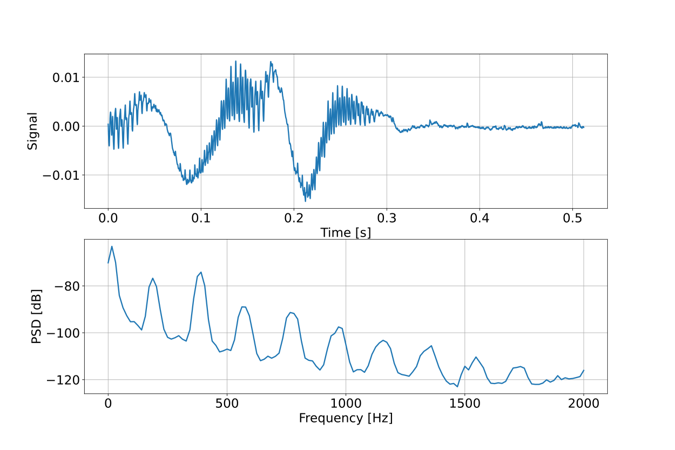}
   \end{center}
   \caption[wbSignals] 
   { \label{fig:wingbeatSignals} Examples of wingbeat signals of two different insect species: (left) Eristalis tenax, (right)  Drosophila drosophila. }
   \end{figure}

The photo diodes are connected in parallel. The signals are pre-amplified using a transimpedance amplifier before being AD converted through a sound card (DELOCK 63926 24bit/ 96kHz), where the signals pass a high pass filter with a cutoff frequency of 8Hz and the DC signal, caused by the insect body, is subtracted. For pre-amplifying we used the OPV TS971 that is suitable for portable devices with a very low noise level of 4ns/$\sqrt{\textrm{Hz}}$ and a high dynamic bandwidth of 12MHz. 
The frequency spectra are Fourier transformed via Welch's method to estimate the spectral density in a Power Spectrum Density (PSD), see Fig.~\ref{fig:wingbeatSignals}. The main frequency so as the higher harmonics indicating significant differences between species can be analysed from the PSD spectra. They are used for the machine learning algorithm. 
Additionally, Spectrograms allow following changes of the wing beat frequencies over time.

\section{Data fusion, ML-Methods and results}
\label{s:datafusion}
As a first proof of concept, we use a small data set of multisensor data and a simple neural network architecture in order to investigate the potential of data fusion of wingbeat and camera data for improving the predictive power.

\subsection{Data set}
\begin{table}[h]
    \centering
\begin{tabular}{|l|l|l|l|l|}
 \toprule
 \textbf{order} & \textbf{family} & \textbf{genus} &  \textbf{species} \\ 
\midrule
\multirow{4}{*}{Hymenoptera} & \multirow{2}{*}{Apidae} & Apis  & mellifera \\ 
 &  &  \multirow{1}{*}{Bombus} & terrestris  \\

     & \multirow{2}{*}{Vespidae} &  Vespa  & crabro \\
   &      & Polistes & dominula  \\    
 \midrule 
Mecoptera & Panorpidae & Panorpa &  communis  \\
 \midrule
 \multirow{2}{*} {Diptera} &  \multirow{2}{*} {Syrphidae} &   Eristalis & tenax  \\
     &    &  Episyrphus  & balteatus \\ 
 \bottomrule
\end{tabular}
\vspace{3mm}
\caption{A small data set of 7 insect species spread over the taxonomic tree is used, see Figure~\ref{fig:histogramm}.}
\label{tab:species}
\end{table}

\begin{figure} [ht]
   \begin{center}
   \includegraphics[width=0.5\textwidth]{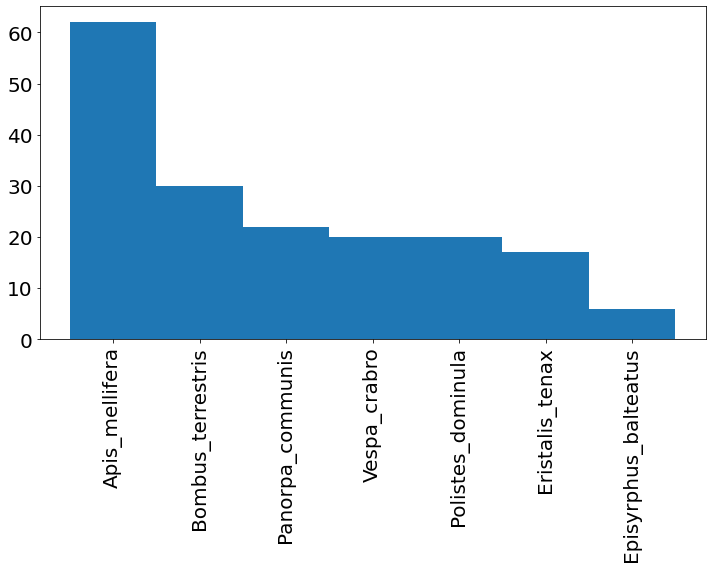} \\
   \includegraphics[width=0.13\textwidth]{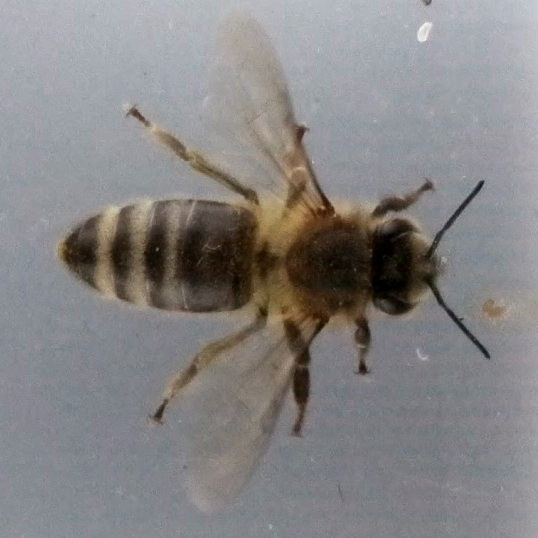}
   \includegraphics[width=0.13\textwidth]{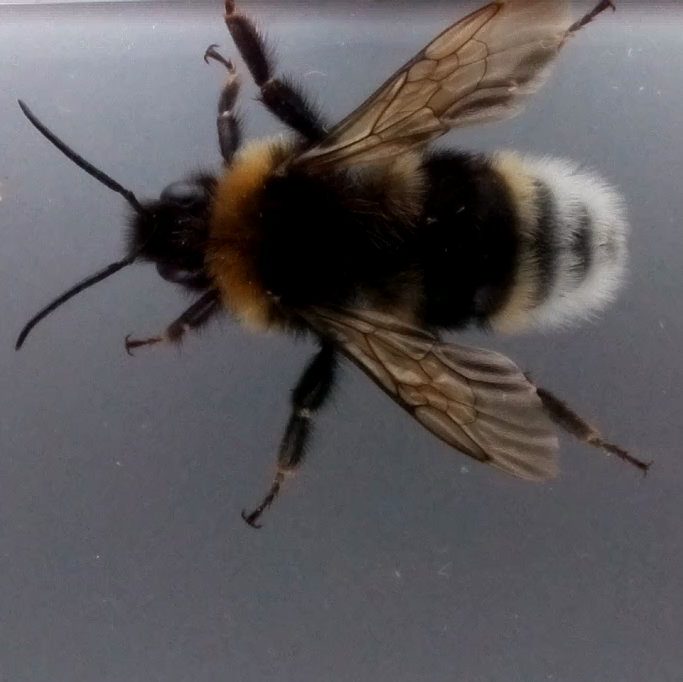}
   \includegraphics[width=0.13\textwidth]{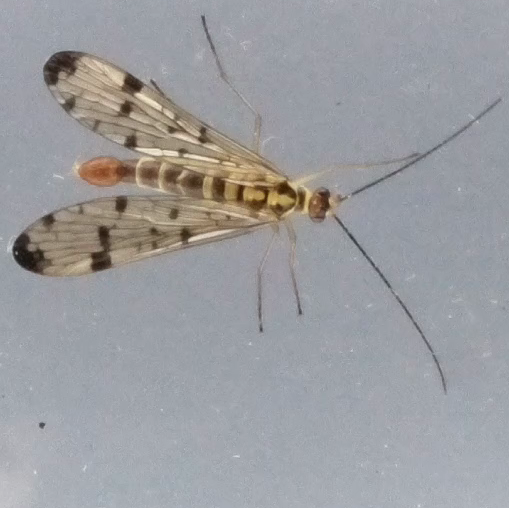}
   \includegraphics[width=0.13\textwidth]{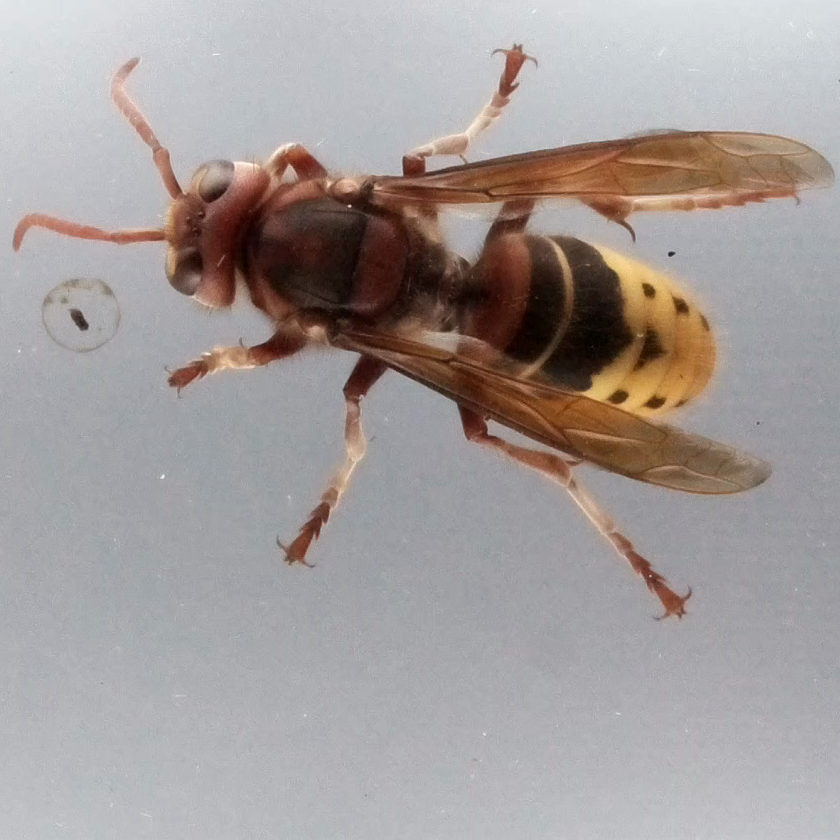}
    \includegraphics[width=0.13\textwidth]{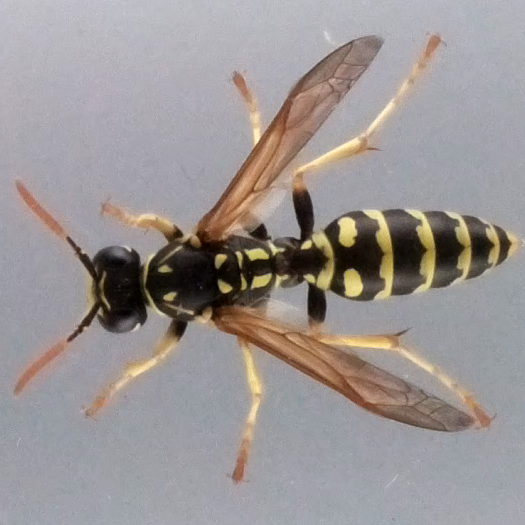}
    \includegraphics[width=0.13\textwidth]{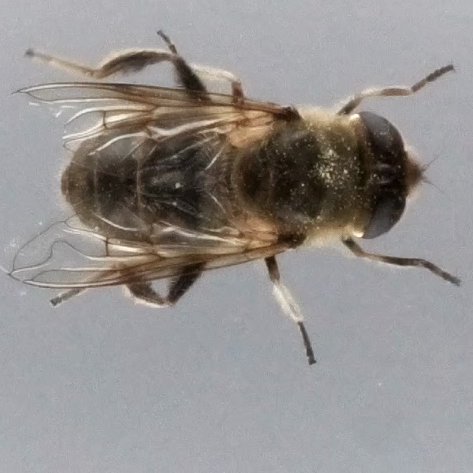}
   \includegraphics[width=0.13\textwidth]{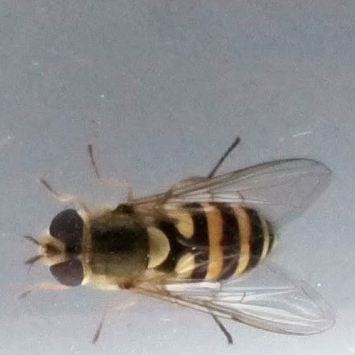}
   \end{center}
   \caption[] 
   { \label{fig:histogramm} A highly unbalanced data set of 7 insect species has been obtained from measurements with the multi sensor system. The presented images have been cropped for better visualisation. Note, that some insects are located at the border of the arena. Also only fast flattering wings show some blurring, but still the image quality is very good.}
   \end{figure}

\subsection{ML-Methods}
Since not all multisensor systems in the field will measure camera and wingbeat signals, we implemented a prediction system, that may deal with three different input data: only camera image, only wingbeat signal or combined camera and wingbeat data.

\begin{figure} [ht]
   \begin{center}
   \begin{tabular}{c} 
   \includegraphics[width=0.8\textwidth]{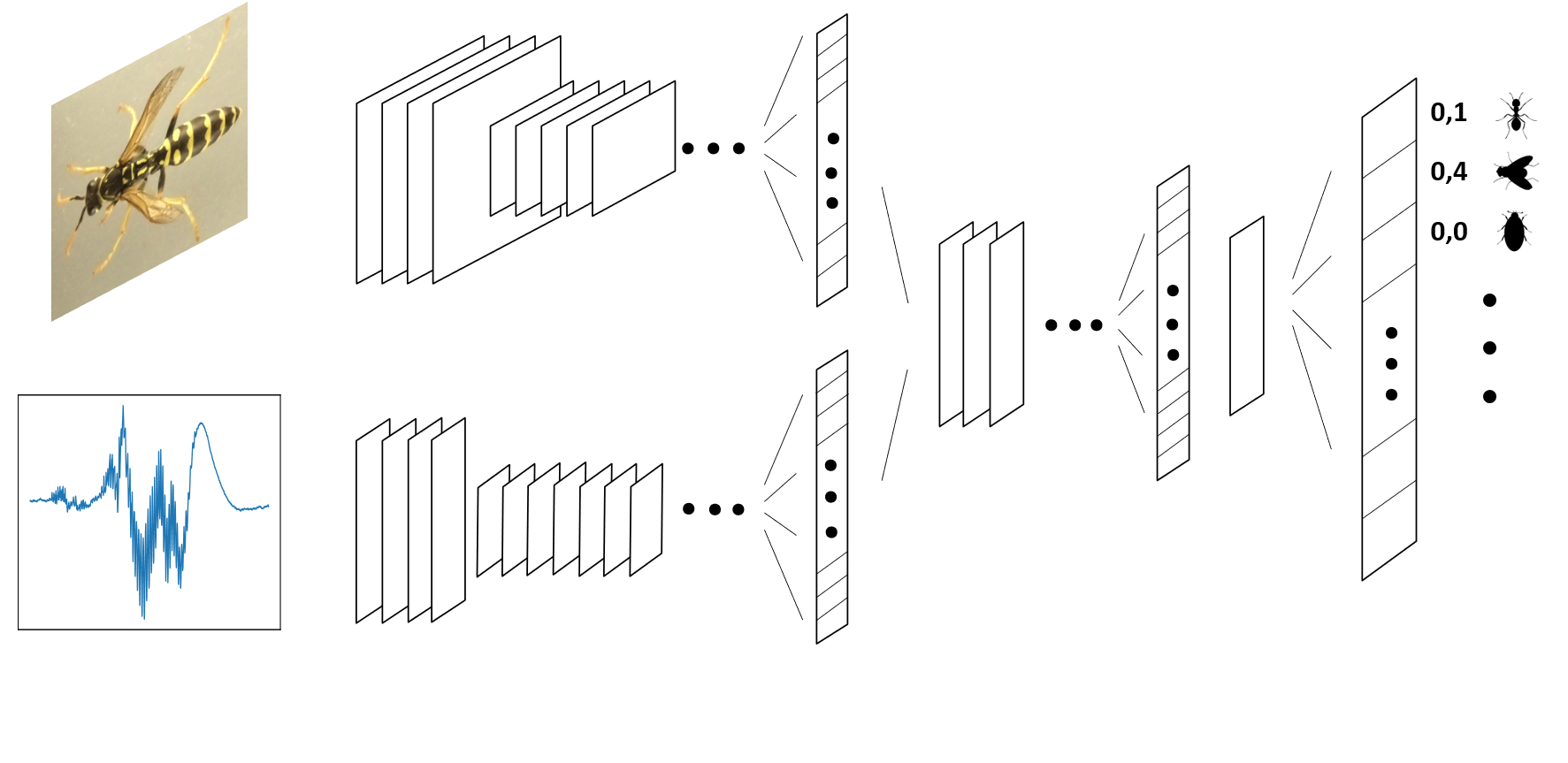}
   \end{tabular}
   \end{center}
   \caption[datafusion] 
   { \label{fig:datafusion} 
   Schematic overview of our neural network architecture for fusion of image and wingbeat data.}
   \end{figure}

Thus, a straight forward approach proceeds as follows: 
\begin{itemize}
    \item[(a)] Implement and train a predictor (NN) for insect classification based on camera images. Use all available camera data for training.
    \item[(b)] Implement and train a predictor (NN) for insect classification based on wing beat signals. Use all available wingbeat signals for training.
    \item[(c)] Use the trained NNs in (a) and (b) as feature extractors and implement and train a predictor on  pairs of camera/wingbeat data.
\end{itemize}
A schematic overview of this architecture is given in Figure~\ref{fig:datafusion}.

At the point of writing this paper, we achieved an overall prediction accuracy of 0.92 on the test set for case (a), using camera images only. Here a mobile net \cite{MobileNet} was trained. For wingbeat signals, case (b), the current data set proved to be to small for training the CNNs as implemented and tested successfully in \cite{teomaster} for larger data sets. However, a linear support vector machine achieved already an accuracy of 0.68.

We are confident to report more results in the near future as soon as more data are available.



\acknowledgments 

Funded by Federal Ministry for the Environment, Nature Conservation, Nuclear Safety and Consumer protection, Project \textit{KInsecta: KI-based Insect Monitoring with Citizen Science}, Funding Nr. 67KI2079.

\bibliography{KInsecta_lib} 

\begin{thebibliography}{10}

\bibitem{Hallmann2017}
Hallmann, C.~A., Sorg, M., Jongejans, E., Siepel, H., Hofland, N., Schwan, H., Stenmans, W., Müller, A., Sumser, H., Hörren, T., Goulson, D., and Kroon, H.~D., ``More than 75 percent decline over 27 years in total flying insect biomass in protected areas,'' {\em PLoS ONE}~{\bf 12} (10 2017).

\bibitem{sanchez2019worldwide}
S{\'a}nchez-Bayo, F. and Wyckhuys, K.~A., ``Worldwide decline of the entomofauna: A review of its drivers,'' {\em Biological conservation}~{\bf 232},  8--27 (2019).

\bibitem{van2018inaturalist}
Van~Horn, G., Mac~Aodha, O., Song, Y., Cui, Y., Sun, C., Shepard, A., Adam, H., Perona, P., and Belongie, S., ``The inaturalist species classification and detection dataset,'' in [{\em Proceedings of the IEEE conference on computer vision and pattern recognition}{\nolinebreak\hspace{0.1em}]},   8769--8778 (2018).

\bibitem{lima2020automatic}
Lima, M. C.~F., de~Almeida~Leandro, M. E.~D., Valero, C., Coronel, L. C.~P., and Bazzo, C. O.~G., ``Automatic detection and monitoring of insect pests—a review,'' {\em Agriculture}~{\bf 10}(5),  161 (2020).

\bibitem{wu2019ip102}
Wu, X., Zhan, C., Lai, Y.-K., Cheng, M.-M., and Yang, J., ``Ip102: A large-scale benchmark dataset for insect pest recognition,'' in [{\em Proceedings of the IEEE/CVF conference on computer vision and pattern recognition}{\nolinebreak\hspace{0.1em}]},   8787--8796 (2019).

\bibitem{vanKlink2022}
van Klink, R., August, T., Bas, Y., Bodesheim, P., Bonn, A., Fossøy, F., Høye, T.~T., Jongejans, E., Menz, M.~H., Miraldo, A., Roslin, T., Roy, H.~E., Ruczyński, I., Schigel, D., Schäffler, L., Sheard, J.~K., Svenningsen, C., Tschan, G.~F., Wäldchen, J., Zizka, V.~M., Åström, J., and Bowler, D.~E., ``Emerging technologies revolutionise insect ecology and monitoring,'' {\em Trends in Ecology and Evolution}~{\bf 37},  872--885 (10 2022).

\bibitem{Mankin2021}
Mankin, R., Hagstrum, D., Guo, M., Eliopoulos, P., and Njoroge, A., ``Automated applications of acoustics for stored product insect detection, monitoring, and management,'' {\em Insects}~{\bf 12} (3 2021).

\bibitem{Rigakis2019}
Rigakis, I., Potamitis, I., Tatlas, N.~A., Livadaras, I., and Ntalampiras, S., ``A multispectral backscattered light recorder of insects’ wingbeats,'' {\em Electronics (Switzerland)}~{\bf 8} (3 2019).

\bibitem{Potamitis2018}
Potamitis, I., Rigakis, I., Vidakis, N., Petousis, M., and Weber, M., ``Affordable bimodal optical sensors to spread the use of automated insect monitoring,'' {\em Journal of Sensors}~{\bf 2018} (2018).

\bibitem{Wang2020}
Wang, J., Zhu, S., Lin, Y., Svanberg, S., and Zhao, G., ``Mosquito counting system based on optical sensing,'' {\em Applied Physics B: Lasers and Optics}~{\bf 126} (2 2020).

\bibitem{Sittinger2023}
{M. Sittinger}, ``Insect detect.'' \url{https://maxsitt.github.io/insect-detect-docs} (2023).
\newblock Online; accessed 30 January 2023.

\bibitem{Brandt2023}
Brandt, D., Tschaikner, M., Chiaburu, T., Schmidt, H., Schrimpf, I., Stadel, A., Beckers, I.~E., and Hau{\ss}er, F., ``Low cost machine vision for insect classification,'' in [{\em Intelligent Systems and Applications}{\nolinebreak\hspace{0.1em}]},  Arai, K., ed.,  18--34, Springer Nature Switzerland, Cham (2024).

\bibitem{Potamitis2016}
Potamitis, I. and Rigakis, I., ``Large aperture optoelectronic devices to record and time-stamp insects’ wingbeats,'' {\em IEEE Sensors Journal}~{\bf 16}(15),  6053--6061 (2016).

\bibitem{MobileNet}
Howard, A.~G., Zhu, M., Chen, B., Kalenichenko, D., Wang, W., Weyand, T., Andreetto, M., and Adam, H., ``Mobilenets: Efficient convolutional neural networks for mobile vision applications,'' (2017).

\bibitem{teomaster}
Chiaburu, T. (2021).

\end{thebibliography}
\bibliographystyle{spiebib} 

\end{document}